\documentclass[twoside]{article}

\usepackage{PRIMEarxiv}
\usepackage{fixltx2e}

\usepackage[utf8]{inputenc} 
\usepackage[T1]{fontenc}    
\usepackage{hyperref}       
\usepackage{url}            
\usepackage{booktabs}       
\usepackage{amsfonts}       
\usepackage{nicefrac}       
\usepackage{microtype}      
\usepackage{lipsum}
\usepackage{fancyhdr}       
\usepackage{graphicx}       
\usepackage{latexsym}       
\usepackage{multirow}
\usepackage[center]{caption}

\usepackage{algpseudocode}
\usepackage{algorithm}

\usepackage{subcaption}

\usepackage{amsthm}
\usepackage{algorithm}
\usepackage{mathtools}  
\usepackage{amsmath}
\usepackage{amssymb}
\usepackage{multirow}
\usepackage{float}
\usepackage{xcolor}
\usepackage{booktabs}
\usepackage{fixltx2e}
\usepackage{adjustbox}
\usepackage{soul}

\usepackage{lineno}

\graphicspath{{images/}}     

\title{OTSU BASED DIFFERENTIAL EVOLUTION METHOD FOR IMAGE SEGMENTATION

}

\author{
  Afreen Shaikh \\
  National Institute of Technology, Warangal \\
  \texttt{safreen@student.nitw.ac.in} \\
  \AND
  Botcha Sharmila \\
  National Institute of Technology, Warangal \\
  \texttt{bsharmila@student.nitw.ac.in} \\
  \And
  Murali Krishna \\
  National Institute of Technology, Warangal \\
  \texttt{mmuralikrishna@student.nitw.ac.in} \\
  \And
  Dr. Sushil Kumar \\
  National Institute of Technology, Warangal \\
  \texttt{kumar.sushil@nitw.ac.in} \\
}

\begin{document}
\maketitle
\begin{abstract}
This paper proposes an OTSU-based differential evolution method for satellite image segmentation and compares it with four other methods such as Modified Artificial Bee Colony Optimizer (MABC)\cite{1}, Artificial Bee Colony (ABC), Genetic Algorithm (GA) and Particle Swarm Optimization (PSO) using the objective function proposed by Otsu for optimal multilevel thresholding. The experiments conducted and their results illustrate that our proposed DE+OTSU algorithm segmentation can effectively and precisely segment the input image, close to results obtained by the other methods. In the proposed DE+OTSU algorithm, instead of passing the fitness function variables, the entire image is passed as an input to the DE algorithm after obtaining the threshold values for the input number of levels in the OTSU's algorithm. The image segmentation results are obtained after learning about the image instead of learning about the fitness variables. In comparison to other segmentation methods examined, the proposed DE+OTSU algorithm yields promising results with minimized computational time comparison to some algorithms.

\end{abstract}


\section{INTRODUCTION}
Image segmentation has a vital part as a pre-processing step in image processing. In particular, it aims at grouping and partitioning pixels within meaningful regions to analyze the image in a detailed manner. Image segmentation applies to varied Computer Vision tasks like Content-based image retrieval, Medical Imaging, Object detection and identification, and Object recognition tasks (like face and fingerprint), among others.

Segmentation methods can be subdivided broadly as (a) edge and line-oriented segmentation methods, (b) region growing methods, (c) clustering, and (d) region splitting methods. Ant Colony Optimisation (ACO), Particle Swarm Optimization (PSO), Bacterial Foraging (BF), and Differential Evolution (DE) are regarded as popular choices for the segmentation of complex images due to their exhaustive searching capability. Because of its clearness and stability, thresholding is coined as one of the most popular image segmentation methods.
Typically, satellite images have insignificant illumination features because of multiple kinds of environmental distributions. They contain diverse objects (regions), like vegetation, water bodies, and territory. However, these areas lack clear demarcation because of low spatial resolution. Hence, segmenting various land cover parts in these images is a complicated task.

One of the key aspects of image segmentation is thresholding (Agrawal et al., 2013)\cite{12}. Many algorithms to perform global level thresholding can be found in the existing literature which aim at segmenting images and extract compelling and meaningful patterns (Rosin, 2001; Portesde et al., 2004; Zahara et al., 2005).

Otsu in the year 1979 \cite{13} presented a method for selecting threshold  from grey-level histograms. But, inadequate articulation of between-class variance raises the cost of computation of the algorithm, specifically in the selection of multi-level threshold. Otsu’s thresholding aims at automatic threshold selection and region-based segmentation and is one of the most accurate techniques for image thresholding because of its simple calculation. Differential Evolution is a reasonably recent population-based evolutionary model which depends on the mutation operation as its prominent step. It efficiently explores large search spaces and exhibits superior results in the following criteria: (1) shorter convergence speed than other evolutionary algorithms, (2) fewer number of parameter adjustments, making it particularly easy to implement.

Tsai (1985)\cite{14} uses the moment-preserving principle to determine thresholds of grey-level input images - the Tsallis entropy technique, a prominent function used for image thresholding. Kittler et al.(1986) work with the assumption that the grey levels of every entity within an image are usually distributed.
Kapur et al. (1985) further introduced a robust method for grey-level image thresholding by utilizing the histogram entropy (Kapur et al., 1985)\cite{15}. 

Entropy-based approaches have earned popularity among all the distinctive image thresholding methods. PCO, influenced by the social behaviours like flocking of birds or schooling of fishes (Akay et al., 2013\cite{16}; Maitra et al. 2008\cite{17}; Yin, et al. 2007\cite{18}), ant colony optimization (ACO) influenced by the pasturing conduct of ant colonies (Ye et al., 2005\cite{20}; Tao et al., 2007\cite{19}) Artificial Bee Colony (ABC) inspired by the pasturing conduct of honey bee swarm (Cuevaset al., 2012 \cite{21}, Akay et al., 2013;).

Another recent area of research inspired by the coordinated intelligence in a swarm of insects or animals has also emerged called Swarm intelligence. One of the widespread and recently developed Swarm Intelligence-based techniques is the ABC algorithm, presented by Karaboga D. (2005)\cite{22}, that simulates the foraging behavior of honey bee colonies. Numerous studies have demonstrated that the performance of Artificial Bee Colony (ABC) method is competitive and comparable with other population-based techniques. 

The MABC method given by Bhandari et al.(2014) introduces an advanced solution search formula (Gao et al., 2012) in charge for its more promising search solution. Within this search equation, the bee explores only closer to the best solution of the iteration before to improve exploitation. The results demonstrate that MABC is best suited for multilevel thresholding of images to attain optimal thresholds for satellite images in comparison to Artificial Bee Colony based algorithms and Particle Swarm Optimization based techniques. Evaluating the advantages, such algorithms are the preferred choice for finding the optimum thresholds in simple images. For example, the Genetic Algorithm (GA) and the improved Genetic Algorithm (Zhang et al., 2014 \cite{23}) are frequently in use to multilevel thresholding scenarios.

The Swarm Intelligence based computing methods can find efficient and optimal solutions for any objective function and have been widely used with an ability to generate highly accurate results in case of complex problems as well. It has also been found by statistical analysis that the Swarm Intelligence based algorithms perform well in multi-level thresholding scenarios (Kurban et al., 2014 \cite{24}).

\section{BRIEF EXPLANATION OF THE ALGORITHMS USED IN THIS STUDY}
\label{sec:headings}

\subsection{Otsu’s thresholding}
\subsubsection{Bi-level thresholding}
In Bi-level thresholding, the aim is to find a threshold to minimize the variance between classes in the segmented image. Otsu's algorithm tends to achieve better results when two different peaks are present in the histogram of the original image, one corresponding to the background and the other to the foreground. 
The entire range of pixels are iterated and the Otsu's threshold is determined when the between-class variances are minimum. Hence, the Otsu's threshold tends to be decided by the class with greater variance.Hence, Otsu's method tends to produce sub-optimal results when there is an occurrence of two or more peaks within the histogram of the image or if one of the classes has a significant variance. 

The total mean and variance are calculated based on the following formulas:
The entire set of pixels are distributed into 2 classes, 

C\textsubscript{1} $\leftarrow$ pixels having grey levels [1, t]\newline
C\textsubscript{2} $\leftarrow$ pixels having grey levels $[t+1, ... ,L]$. \newline
where $t$ corresponds to Otsu's threshold

The probability distribution of the two classes is denoted by: 
\begin{equation}
C_{1}:{\frac {p_{1}} {w_{1}(t)}, ..., \frac {p_{r}} {w_{1}(t)}} 
\end{equation}
\begin{center}
    and
\end{center}
\begin{equation}
C_{2}:{\frac {p_{r+1}} {w_{2}(t)}, ..., \frac {p_{L}} {w_{2}(t)}} 
\end{equation}

Where, $w_{1}(t) = \sum _{i=1}^{r}p_{i}$ and $w_{2}(t) = \sum _{i=r+1}^{L}p_{i}$

The class mean for the two classes $\mu_{1}$ and $\mu_{2}$ are defined by: 
\begin{equation}
\mu_{1}=\sum _{i=1}^{r} \frac{i*p_{i}}{w_{1}(t)}
\end{equation}

\begin{equation}
\mu_{2}=\sum _{i=r+1}^{L} \frac{i*p_{i}}{w_{2}(t)}
\end{equation}

Otsu's between-class variance based on discriminate analysis of the threshold image is defined as:
\begin{equation}
\sigma^2_{B} = w_{1}(\mu_{1}-\mu_{T})^2 + w_{2}(\mu_{2}-\mu_{T})^2
\end{equation}

For bi-level thresholding, the optimal threshold t* is chosen to maximize between-class variance n, i.e.
\begin{equation}
t^* = arg_{t<l}max{\sigma^2_{B}(t)}
\end{equation}

\subsubsection{Multilevel Thresholding}
The first step in the algorithm is to obtain the Otsu's threshold and the class means of the two classes divided by the Otsu's threshold. Further, in multi-level thresholding, the pixels of the image are divided into three categories instead of two categories determined by Otsu's threshold. The three categories correspond to the following: (a) the 'foreground' region - group of pixel with values >= the larger mean, (b) the 'background' region - group of pixels with values <= to the smaller mean and (c) the 'to-be-determined (TBD)' region - group of pixels with values between the two class means.

In iteration $i+1$, the algorithm retains the 'foreground' and 'background' regions from iteration $i$ and re-applies the Otsu's method only on the 'to-be-determined' region to further divide it into three classes again. When this iteration stops after satisfying a pre-defined criterion, the final 'to-be-determined (TBD)' region is then divided into two classes - foreground and background instead of three. Lastly, the foreground regions from all the iterations are combined to get the final foreground class and the final background region is also determined likewise.

For an image represented by $L$ no.of grey levels $0, 1, . . ., L - 1$, we could develop the image histogram $H = \{f_{0},f_{1}, . . ., f_{L-1}\}$, where $f_{i}$ is the frequency of grey level $i$ in the image. Let  \nolinebreak\[N = \sum_{i=0}^{L-1}f_{i}\] determine the total no. of pixels in the image. The occurrence probability of $i^{th}$ grey level is defined by:

\begin{equation}
p_{i} = \frac{f_{i}}{N}
\end{equation}

It can be easily be illustrated that $p_{i} \ge 0$ and  $\sum_{i=0}^{L-1}p_{i}=1$. 

Otsu’s algorithm divides the image into $K + 1$ clusters $\{C0,C1, . . .,C_{K}\}$ using $K$ no. of thresholds chosen from the set $T = \{(t_{1},t_{2}, . . .,t_{K} ) |  0<t1< . . . < t_{K}< L\}$ where $C_{K}$ is the set of pixels with grey levels $\{t_{K}, t_{K}+1, ... ,t_{K+1} - 1\}$. Where $t_{0} = 0$ and $t_{K+1} = L$. For every cluster $C_{K}$, the cumulative probability $w_{K}$ and mean grey level $\mu_{K}$ are defined by:

\begin{equation}
w_{k} = \sum_{i \epsilon C_{k}} , k \epsilon \{0,1,2,...,K\}
\end{equation}

\begin{equation}
\mu_{k} = \sum_{i \epsilon C_{k}} \frac{i*p_{i}}{w_{k}} , k \epsilon \{0,1,2,...,K\}
\end{equation}

The mean intensity of the whole image $\mu_{T}$ and the between-class variance $\sigma_{B}^2$ are defined by:
\begin{equation}
\mu_{T} = \sum_{k=0}^K w_{k} * \mu_{k} = \sum_{i=0}^{L-1} i*p_{i}
\end{equation}
\begin{center}
    and
\end{center}

\begin{equation}
\sigma_{B}^2 = \sum_{k=0}^{K} w_{k} * (\mu_{k} - \mu_{T})^2 = \sum_{k=0}^K w_{k} * \mu_{k}^2 - \mu_{T}^2
\end{equation}

In Otsu’s method, the threshold levels for each cluster are selected on the basis of maximizing the variance among means of the cluster (Huang et al., 2011). The optimal thresholds at maximum between-class variance are determined by: 
\begin{equation}
(t_{1}^*,t_{2}^*,...,t_{K}^*) = \{\sigma_{B}^2(t_{1}, t_{2},..., t_{K})\}
\end{equation}

\section{DIFFERENTIAL EVOLUTION (DE)}
\label{sec:DE}
In comparison to other evolutionary algorithms, the Differential Evolution (DE) algorithm aims to reserve the global search strategy based on population and employs a simple mutation function of the differential and one-on-one competition, reducing the operation's genetic complexity. At the same time, the specific memory ability of DE allows it to dynamically track the current search to adjust its search strategy with robust global convergence. Hence, it is suited for the complex environments of the optimization problem. Fundamental operations like selection, crossover, and mutation are the foundation of the Differential Evolution algorithm. 

\subsection{How DE is used in the proposed algorithm?}
Initially, the image to be segmented is made to run through OTSU’s multilevel thresholding, on the basis of the number of levels($L$) the image has to be segmented. The output of this step yields us with ‘$L$’ threshold values which act as partitions for the pixel intensities. These partitions divide the pixel intensity range (0-255) into ‘$L+1$’ clusters. Based on the pixels belonging to the image that fall into a particular cluster, the cluster centre is defined using the mean of the pixels (falling in that cluster).

Now, the image to be segmented is given as input to the DE algorithm. Each of the ‘$G$’ generations has Np number of populations, where population size is equal to the count of the image pixels. After each generation, the best population is evaluated based on the least value of fitness function, which is a measure to the MSE (mean square error) distance of each pixel in the population to its corresponding cluster centre. The next generations are populated using this best population as described later in the algorithm. The best population of the last generation is again reshaped into image format, which form the segmented image.

\section{PROPOSED ALGORITHM}
\subsection{Algorithm pre-processing}
In our implementation an input image and the number of levels ($L$) are given as input to
the algorithm which would return the cluster partition values $(cp_{1}, cp_{2}, ... , cp_{L})$.

Cluster ranges, $(0, cp_{1}), (cp_{1}+1, cp_{2}),..., (cp_{L}+1, 255)$ are computed from the cluster partition values obtained after the thresholding of the image.

Cluster centres, $(cc_{1} , cc_{2}, ... , cc_{L+1})$ are calculated as follows :

\begin{equation}
cc_{i} = \frac{\sum pix^i}{|pix^i|}
\end{equation}

Where,

$pix^i$ = pixels in the cluster range $(0, cp_{i})$

$|pix^i|$ = pixels in the cluster range $(0, cp_{i})$

\subsection{Differential Evolution}

Consider the image has ‘$m$’ pixels $(pix_{1}, pix_{2}, . . , pix_{m})$, and algorithm runs for ‘$G$’
generations with ‘$Np$’ no. of populations per generation.

\begin{center}
    $X_{n,i}^g$ : $i^{th}$ pixel value of $n^{th}$ population in $g^{th}$ generation
\end{center}

\begin{equation}
X_{n}^g : [X_{n,1}^g, X_{n,2}^g, ..., X_{n,m}^g]
\end{equation}

\begin{center}
$n: 1 \rightarrow Np (population index)$
    
$g: 1 \rightarrow G (generation index)$
    
$i: 1 \rightarrow m (pixel index)$
    
\end{center}

\begin{algorithm}[hbt!]
\caption{An algorithm with caption}\label{alg:cap}
\begin{algorithmic}
\State $n = 0$
\State $i = 0$
\While{$n \neq 0$}
    \While{$i \neq P$}
        \State $[Xlow, Xhigh] \leftarrow cluster range for pixel, i$
        \State $X^1_{n,i} \leftarrow X_{low} + rand(0,1)*[X_{high} - X_{low}]$
    \EndWhile
\EndWhile
\State $n = 0$
\While{$n \neq 0$}
    \State $MSE_{X^{g=1}_{n}} = \frac{1}{n}\sum_{i=1}^m(X_{n,i}^1-C_{i})^2$
    \State $X_{best}^1 \leftarrow population with minimum fitness value$
\EndWhile\newline
Mutant vector :\newline
for each population n in generation g+1
\begin{algorithmic}
\State $V_{n}^{g+1} \leftarrow X_{best}^g + F*(X_{r1}^g-X_{r2}^g)$
    \If{$rand(0,1)<=Cr$}
        \State $U_{n}^{g+1} \leftarrow V_{n}^{g+1}$
    \ElsIf{$rand(0,1)>Cr$}
        \State $U_{n}^{g+1} \leftarrow X_{n}^g$
    \EndIf
    \If{$fitness(U_{n}^{g+1} < fitness(X_{n}^{g})$}
        \State $X_{n}^{g+1} \leftarrow U_{n}^{g+1}$
    \ElsIf{$fitness(U_{n}^{g+1} > fitness(X_{n}^{g})$}
        \State $X_{n}^{g+1} \leftarrow X_{n}^{g}$
    \EndIf
\end{algorithmic}


Fitness function and mutant vector calculation will be repeated for G generations.
After G generations the $X_{best}^G$ population is considered the solution to image segmentation and the ‘m’ pixels in $X_{best}^G$ is converted into output image

\end{algorithmic}
\end{algorithm}


\section{RESULTS}
\label{sec:results}
\subsection{Performance Metrics}
This section aims at presenting results gathered from various satellite images using the proposed OTSU based differential evolution algorithm are discussed and those results are set side by side the results obtained from the Modified Artificial Bee Colony algorithm (MABC) \cite{11}, Artificial Bee Colony algorithm (ABC), Particle Swarm Optimization algorithm (PSO) and Genetic Algorithm (GA). Along with quality estimation factor of Peak Signal-to-Noise Ratio (PSNR) ,algorithm efficiency (based on CPU Timing) and feature assessment is also measured using Structural Similarity Index measure (SSIM).

Table 6-14 compares the PSNR, SSIM and CPU time values gathered by applying the proposed DE+OTSU method and set side by side for comparison with the result obtained using Modified Artificial Bee Colony algorithm (MABC), Artificial Bee Colony algorithm (ABC), Particle Swarm Optimization algorithm (PSO) and Genetic Algorithm (GA) methods respectively.

\subsection{Input Images and References}

\begin{figure}[!htb]

\begin{subfigure}[b]{0.3\textwidth}
    \includegraphics[width=\textwidth]{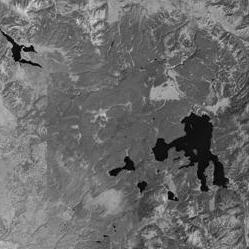}
    \caption{}
    \label{fig:first}
\end{subfigure}
\hfill
\begin{subfigure}[b]{0.3\textwidth}
    \includegraphics[width=\textwidth]{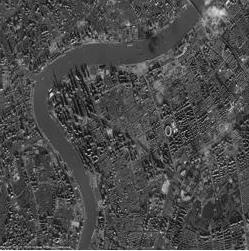}
    \caption{}
    \label{fig:second}
\end{subfigure}
\hfill
\begin{subfigure}[b]{0.3\textwidth}
    \includegraphics[width=\textwidth]{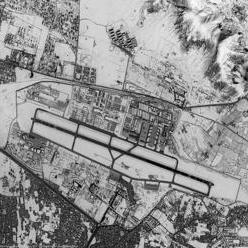}
    \caption{}
    \label{fig:third}
\end{subfigure}
\caption{Satellite images used to perform and compare the experiments\newline \centering (a) an image from the NASA Earth Observatory, Yellowstone National Park, June 2009 \centering (b)Pléiades Satellite Image: Shanghai, China \centering (c)Pléiades Satellite Image: Kabul Airport, Afghanistan
}
\label{fig:SatelliteImgs}
\end{figure}

\subsection{Parameters configuration in the Proposed Algorithm}
\begin{table}[!htb]
 \caption{Parameters used in DE+OTSU}
  \centering
  \begin{tabular}{lll}
  \toprule
    Parameters     & Value     \\
    \midrule
    No. of generations and population size &  10 and 100     \\
    Crossover and mutation probability     & 0.2, 0.3      \\
    Image size     & 250*250 pixels  \\
    \bottomrule
  \end{tabular}
  \label{tab:table}
\end{table}

\begin{table}[!htb]
 \caption{Parameters configuration in ABC}
  \centering
  \begin{tabular}{lll}
  \toprule
    Parameters     & Value     \\
    \midrule
    Swam Size and iterations &  10 and 100     \\
    Lower and Upper bounds $(lb (X_{min}) , ub (X_{max}))$     & 1 and 256      \\
    Trial limit at max     & 10  \\
    $F_{i}(\phi)$ & [0,1]  \\
    \bottomrule
  \end{tabular}
  \label{tab:table}
\end{table}

\begin{table}[!htb]
 \caption{Parameters configuration in MABC}
  \centering
  \begin{tabular}{lll}
  \toprule
    Parameters     & Value     \\
    \midrule
    Swam Size &  10     \\
    Iterations (Max FE)     & 100      \\
    No. of Iterations with chaos     & 300  \\
    Lower and Upper bounds $(lb (X_{min}) , ub (X_{max}))$ & 1 and 256  \\
    \bottomrule
  \end{tabular}
  \label{tab:table}
\end{table}

\begin{table}[!htb]
 \caption{Parameters configuration in PSO}
  \centering
  \begin{tabular}{lll}
  \toprule
    Parameters     & Value     \\
    \midrule
    Swam Size and iterations &  200 and 100     \\
    Cognitive, social and neighbourhood acceleration (C1, C2, C3)     & 2,2,1      \\
    Lower and Upper bounds $(lb (X_{min}) , ub (X_{max}))$     & 1 and 256  \\
    Error goal and Trial limit at max & $1e^-7$ and 500  \\
    Starting velocity weight & 0.95  \\
    Ending velocity weight & 0.4  \\
    The proportion of iterations, for linear varied W & 0.7 \\
    Constriction Factor, Maximum Velocity Step and Neighborhood Size & 1  \\
    Global Minima & 0  \\
    \bottomrule
  \end{tabular}
  \label{tab:table}
\end{table}

\begin{table}[!htb]
 \caption{Parameters configuration in GA}
  \centering
  \begin{tabular}{lll}
  \toprule
    Parameters     & Value     \\
    \midrule
    Size of the population and iterations & 10 and 100 \\
Crossover and mutation probability & 0.5 and 0.1 \\
Bits used in variable representation & [8] \\
$eta (\eta)$ & 1 \\
Lower and Upper  bounds $(lb (X_{min}) , ub (X_{max}))$ & 1 and 256 \\
    \bottomrule
  \end{tabular}
  \label{tab:table}
\end{table}
\subsection{Comparison of Output Images}

Table 6 : Comparison between Modified Artificial Bee Colony algorithm (MABC), Artificial Bee Colony algorithm (ABC), Particle Swarm Optimization algorithm (PSO) and Genetic Algorithm (GA) algorithms using Kapur’s entropy with proposed DE+OTSU based on CPU timing (seconds).
$$
\begin{array}{|c|c|l|c|c|c|c|}
\hline \multirow{2}{*}{\text { Image }} & \multirow{2}{*}{\text { Level }} & \multirow{2}{*}{\text { DE+OTSU }} & \multicolumn{4}{|c|}{\text { KAPUR'S ENTROPY }} \\
\cline { 4 - 7 } & & & \text { MABC } & \text { ABC } & \text { PSO } & \text { GA } \\
\hline \multirow{4}{*}{1} & 2 & 10.62 & 2.843121 & 6.890611 & 21.00694 & 185.94096 \\
\cline { 2 - 7 } & 3 & 10.96 & 2.48574 & 6.641995 & 22.24080 & 181.80301 \\
\cline { 2 - 7 } & 4 & 8.26 & 2.528806 & 6.636246 & 21.84823 & 184.04964 \\
\cline { 2 - 7 } & 5 & 9.92 & 3.181302 & 6.661542 & 23.07188 & 121.46275 \\
\hline \multirow{4}{*}{2} & 2 & 12.57 & 2.947659 & 6.510813 & 24.00778 & 268.01849 \\
\cline { 2 - 7 } & 3 & 10.56 & 2.487444 & 6.295622 & 23.27194 & 266.77671 \\
\cline { 2 - 7 } & 4 & 8.69 & 2.621718 & 7.009322 & 23.85787 & 268.79580 \\
\cline { 2 - 7 } & 5 & 10.65 & 2.733291 & 6.570533 & 24.95336 & 183.32426 \\
\hline \multirow{4}{*}{3} & 2 & 9.93 & 2.81911 & 8.235545 & 25.22988 & 436.11369 \\
\cline { 2 - 7 } & 3 & 11.36 & 3.286868 & 8.075897 & 24.80887 & 433.87366 \\
\cline { 2 - 7 } & 4 & 8.11 & 2.612452 & 6.917417 & 24.86667 & 438.91758 \\
\cline { 2 - 7 } & 5 & 10.34 & 2.821316 & 8.247721 & 25.73357 & 296.15088 \\
\hline
\end{array}
$$

\newpage

Table 7 : Comparison between Modified Artificial Bee Colony algorithm (MABC), Artificial Bee Colony algorithm (ABC), Particle Swarm Optimization algorithm (PSO) and Genetic Algorithm (GA) algorithms using Kapur’s entropy with proposed DE+OTSU based on PSNR (dB) \cite{25} values.
$$
\begin{array}{|c|l|l|l|l|l|l|}
\hline \multirow{2}{*}{\text { Image }} & \multirow{2}{*}{\text { Level }} & \multirow{2}{*}{\text { DE+OTSU }} & \multicolumn{4}{|c|}{\text { KAPUR'S ENTROPY }} \\
\cline { 4 - 7 } & & & \text { MABC } & \text { ABC } & \text { PSO } & \text { GA } \\
\hline \multirow{6}{*}{1} & 2 & 16.43 & 24.54 & 24.53 & 24.55 & 24.58 \\
\cline { 2 - 7 } & 3 & 15.53 & 24.76 & 24.79 & 24.74 & 24.756 \\
\cline { 2 - 7 } & 4 & 20.16 & 24.97 & 25.01 & 24.969 & 25.89 \\
\cline { 2 - 7 } & 5 & 18.94 & 25.82 & 25.76 & 25.188 & 25.676 \\
\hline \multirow{4}{*}{2} & 2 & 16.77 & 24.6265 & 24.606 & 24.575 & 24.591 \\
\cline { 2 - 7 } & 3 & 18.85 & 24.80 & 24.801 & 24.84 & 24.811 \\
\cline { 2 - 7 } & 4 & 20.12 & 25.024 & 25.008 & 25.008 & 25.979 \\
\cline { 2 - 7 } & 5 & 21.71 & 25.542 & 25.238 & 26.106 & 25.54 \\
\hline \multirow{3}{*}{3} & 2 & 14.36 & 24.64 & 24.659 & 24.649 & 24.691 \\
\cline { 2 - 7 } & 3 & 19.53 & 24.83 & 24.86 & 24.84 & 24.90 \\
\cline { 2 - 7 } & 4 & 20.99 & 25.065 & 25.039 & 25.063 & 25.978 \\
\cline { 2 - 7 } & 5 & 20.54 & 24.5451 & 24.542 & 24.551 & 24.582 \\
\hline
\end{array}
$$

Table 8 : Comparison between Modified Artificial Bee Colony algorithm (MABC), Artificial Bee Colony algorithm (ABC), Particle Swarm Optimization algorithm (PSO) and Genetic Algorithm (GA) algorithms using Kapur’s entropy with proposed DE+OTSU based on SSIM \cite{25} values.
$$
\begin{array}{|c|c|l|c|c|c|c|}
\hline \multirow{2}{*}{\text { Image }} & \multirow{2}{*}{\text { Level }} & \multirow{2}{*}{\text { DE+OTSU }} & \multicolumn{4}{|c|}{\text { KAPUR’S ENTROPY }} \\
\cline { 4 - 7 } & & & \text { MABC } & \text { ABC } & \text { PSO } & \text { GA } \\
\hline \multirow{6}{*}{1} & 2 & 0.37 & 0.900747 & 0.865230 & 0.936475 & 0.941127 \\
\cline { 2 - 7 } & 3 & 0.312 & 0.966901 & 0.953207 & 0.965857 & 0.962966 \\
\cline { 2 - 7 } & 4 & 0.59 & 0.97696 & 0.970332 & 0.977355 & 0.969124 \\
\cline { 2 - 7 } & 5 & 0.492 & 0.989512 & 0.981282 & 0.984123 & 0.979080 \\
\hline \multirow{4}{*}{2} & 2 & 0.53 & 0.886746 & 0.877771 & 0.945456 & 0.945627 \\
\cline { 2 - 7 } & 3 & 0.59 & 0.958891 & 0.949243 & 0.967704 & 0.965602 \\
\cline { 2 - 7 } & 4 & 0.761 & 0.976986 & 0.974446 & 0.979841 & 0.975580 \\
\cline { 2 - 7 } & 5 & 0.672 & 0.985457 & 0.978247 & 0.985389 & 0.982706 \\
\hline \multirow{3}{*}{3} & 2 & 0.421 & 0.918842 & 0.902112 & 0.943819 & 0.947394 \\
\cline { 2 - 7 } & 3 & 0.672 & 0.961164 & 0.959592 & 0.968163 & 0.963107 \\
\cline { 2 - 7 } & 4 & 0.73 & 0.978343 & 0.961763 & 0.979184 & 0.976280 \\
\cline { 2 - 7 } & 5 & 0.704 & 0.989280 & 0.977779 & 0.985351 & 0.981165 \\
\hline
\end{array}
$$

Table 9 : Comparison between Modified Artificial Bee Colony algorithm (MABC), Artificial Bee Colony algorithm (ABC), Particle Swarm Optimization algorithm (PSO) and Genetic Algorithm (GA) algorithms using Between-class variance with proposed DE+OTSU based on CPU timing (seconds).

$$
\begin{array}{|c|c|l|c|c|c|c|}
\hline \multirow{2}{*}{\text { Image }} & \multirow{2}{*}{\text { Level }} & \multirow{2}{*}{\text { DE+OTSU }} & \multicolumn{4}{|c|}{\text { BETWEEN CLASS VARIANCE }} \\
\cline { 4 - 7 } & & & \text { MABC } & \text { ABC } & \text { PSO } & \text { GA } \\
\hline \multirow{4}{*}{1} & 2 & 10.62 & 2.251826 & 2.511814 & 13.23507 & 121.92735 \\
\cline { 2 - 7 } & 3 & 10.96 & 2.383504 & 2.495365 & 4.30062 & 118.78825 \\
\cline { 2 - 7 } & 4 & 8.26 & 2.243267 & 2.596011 & 4.000213 & 119.73264 \\
\cline { 2 - 7 } & 5 & 9.92 & 2.228959 & 2.434094 & 4.336584 & 119.81306 \\
\hline \multirow{4}{*}{2} & 2 & 12.57 & 3.244759 & 3.504219 & 5.748055 & 174.33693 \\
\cline { 2 - 7 } & 3 & 10.56 & 3.437252 & 3.588553 & 5.722739 & 174.18969 \\
\cline { 2 - 7 } & 4 & 8.69 & 3.440058 & 3.744837 & 5.670794 & 174.57630 \\
\cline { 2 - 7 } & 5 & 10.65 & 3.406708 & 3.732828 & 5.793264 & 173.69577 \\
\hline \multirow{4}{*}{3} & 2 & 9.93 & 4.667568 & 5.007778 & 7.29691 & 272.89039 \\
\cline { 2 - 7 } & 3 & 11.36 & 5.363554 & 5.276051 & 6.587852 & 275.03171 \\
\cline { 2 - 7 } & 4 & 8.11 & 5.076405 & 5.519536 & 7.154403 & 272.78504 \\
\cline { 2 - 7 } & 5 & 10.34 & 4.784902 & 5.427858 & 6.753919 & 275.44405 \\
\hline
\end{array}
$$

Table 10 : Comparison between Modified Artificial Bee Colony algorithm (MABC), Artificial Bee Colony algorithm (ABC), Particle Swarm Optimization algorithm (PSO) and Genetic Algorithm (GA) algorithms using Between-class variance with proposed DE+OTSU based on PSNR (dB) \cite{25} values.
$$
\begin{array}{|c|c|c|c|c|c|c|}
\hline \multirow{2}{*}{\text { Image }} & \multirow{2}{*}{\text { Level }} & \multirow{2}{*}{\text { DE+OTSU }} & \multicolumn{4}{|c|}{\text { BETWEEN CLASS VARIANCE }} \\
\cline { 4 - 7 } & & & \text { MABC } & \text { ABC } & \text { PSO } & \text { GA } \\
\hline \multirow{3}{*}{1} & 2 & 16.43 & 24.56352 & 24.579535 & 24.546290 & 24.552551 \\
\cline { 2 - 7 } & 3 & 15.53 & 24.780481 & 24.759637 & 24.732067 & 24.764200 \\
\cline { 2 - 7 } & 4 & 20.16 & 24.981269 & 25.006653 & 25.012021 & 24.874078 \\
\cline { 2 - 7 } & 5 & 18.94 & 25.627989 & 25.4729200 & 25.197580 & 25.004700 \\
\hline \multirow{4}{*}{2} & 2 & 16.77 & 24.621500 & 24.583300 & 24.585789 & 24.570418 \\
\cline { 2 - 7 } & 3 & 18.85 & 24.800300 & 24.794900 & 24.836735 & 24.786600 \\
\cline { 2 - 7 } & 4 & 20.12 & 24.968400 & 25.004600 & 24.973612 & 24.974895 \\
\cline { 2 - 7 } & 5 & 21.71 & 25.771400 & 25.300500 & 25.234673 & 25.158200 \\
\hline \multirow{3}{*}{3} & 2 & 14.36 & 24.653841 & 24.648178 & 24.644845 & 24.645697 \\
\cline { 2 - 7 } & 3 & 19.53 & 24.856153 & 24.870026 & 24.859707 & 24.870600 \\
\cline { 2 - 7 } & 4 & 20.99 & 25.067745 & 25.054861 & 25.066102 & 26.279928 \\
\cline { 2 - 7 } & 5 & 20.54 & 25.769200 & 25.082472 & 25.293798 & 25.124000 \\
\hline
\end{array}
$$

Table 11 : Comparison between Modified Artificial Bee Colony algorithm (MABC), Artificial Bee Colony algorithm (ABC), Particle Swarm Optimization algorithm (PSO) and Genetic Algorithm (GA) algorithms using Between-class variance with proposed DE+OTSU based on SSIM \cite{25} values.
$$
\begin{array}{|c|c|c|c|c|c|c|}
\hline \multirow{2}{*}{\text { Image }} & \multirow{2}{*}{\text { Level }} & \multirow{2}{*}{\text { DE+OTSU }} & \multicolumn{4}{|c|}{\text { BETWEEN CLASS VARIANCE }} \\
\cline { 4 - 7 } & & & \text { MABC } & \text { ABC } & \text { PSO } & \text { GA } \\
\hline \multirow{4}{*}{1} & 2 & 0.37 & 0.907199 & 0.898708 & 0.9341685 & 0.936426 \\
\cline { 2 - 7 } & 3 & 0.312 & 0.958494 & 0.950182 & 0.9613643 & 0.964400 \\
\cline { 2 - 7 } & 4 & 0.59 & 0.97805 & 0.976883 & 0.9771952 & 0.924582 \\
\cline { 2 - 7 } & 5 & 0.492 & 0.984679 & 0.978082 & 0.9837941 & 0.972000 \\
\hline \multirow{4}{*}{2} & 2 & 0.53 & 0.885600 & 0.869500 & 0.9450091 & 0.995003 \\
\cline { 2 - 7 } & 3 & 0.59 & 0.958300 & 0.957800 & 0.9666987 & 0.962600 \\
\cline { 2 - 7 } & 4 & 0.761 & 0.975500 & 0.978500 & 0.9784969 & 0.960120 \\
\cline { 2 - 7 } & 5 & 0.672 & 0.981600 & 0.979800 & 0.9806148 & 0.947300 \\
\hline \multirow{4}{*}{3} & 2 & 0.421 & 0.920762 & 0.904131 & 0.9423958 & 0.942725 \\
\cline { 2 - 7 } & 3 & 0.672 & 0.963812 & 0.968801 & 0.9687778 & 0.962500 \\
\cline { 2 - 7 } & 4 & 0.73 & 0.971162 & 0.972597 & 0.9778467 & 0.955299 \\
\cline { 2 - 7 } & 5 & 0.704 & 0.982051 & 0.967690 & 0.9819275 & 0.95570000 \\
\hline
\end{array}
$$

Table 12 : Comparison between Modified Artificial Bee Colony algorithm (MABC), Artificial Bee Colony algorithm (ABC), Particle Swarm Optimization algorithm (PSO) and Genetic Algorithm (GA) algorithms using Tsallis entropy with proposed DE+OTSU based on CPU timing (seconds).
$$
\begin{array}{|l|l|l|l|l|c|c|}
\hline \multirow{2}{*}{\text { Image }} & \multirow{2}{*}{\text { Level }} & \multirow{2}{*}{\text { DE+OTSU }} & \multicolumn{4}{|c|}{\text { TSALLIS ENTROPY }} \\
\cline { 4 - 7 } & & & \text { MABC } & \multicolumn{1}{|c|}{\text { ABC }} & \text { PSO } & \text { GA } \\
\hline \multirow{4}{*}{1} & 2 & 10.62 & 4.191848 & 9.2953 & 5.285757 & 1287.455 \\
\cline { 2 - 7 } & 3 & 10.96 & 4.28949 & 10.4439 & 5.165659 & 1243.575 \\
\cline { 2 - 7 } & 4 & 8.26 & 5.003336 & 9.4859 & 5.261193 & 0129.445 \\
\cline { 2 - 7 } & 5 & 9.92 & 5.064827 & 8.9783 & 5.146160 & 0122.208 \\
\hline \multirow{4}{*}{2} & 2 & 12.57 & 4.687289 & 9.9036 & 7.547844 & 1804.159 \\
\cline { 2 - 7 } & 3 & 10.56 & 5.839503 & 8.8693 & 6.745230 & 1813.836 \\
\cline { 2 - 7 } & 4 & 8.69 & 5.344366 & 9.3367 & 6.553589 & 0178.122 \\
\cline { 2 - 7 } & 5 & 10.65 & 6.290270 & 9.0729 & 7.058079 & 0180.25 \\
\hline \multirow{4}{*}{3} & 2 & 9.93 & 6.721415 & 8.9362 & 7.839406 & 2672.361 \\
\cline { 2 - 7 } & 3 & 11.36 & 6.710296 & 10.7472 & 7.785332 & 2644.285 \\
\cline { 2 - 7 } & 4 & 8.11 & 6.903615 & 10.5427 & 7.210223 & 297.826 \\
\cline { 2 - 7 } & 5 & 10.34 & 7.755440 & 10.1722 & 7.334682 & 263.15 \\
\hline
\end{array}
$$

Table 13 : Comparison between Modified Artificial Bee Colony algorithm (MABC), Artificial Bee Colony algorithm (ABC), Particle Swarm Optimization algorithm (PSO) and Genetic Algorithm (GA) algorithms using Tsallis entropy with proposed DE+OTSU based on PSNR (dB) \cite{25} values.
$$
\begin{array}{|l|l|l|c|c|c|c|}
\hline \multirow{2}{*}{\text { Image }} & \multirow{2}{*}{\text { Level }} & \multirow{2}{*}{\text { DE+OTSU }} & \multicolumn{4}{|c|}{\text { TSALLIS ENTROPY }} \\
\cline { 4 - 7 } & & & \text { MABC } & \text { ABC } & \text { PSO } & \text { GA } \\
\hline \multirow{3}{*}{1} & 2 & 16.43 & 24.377843 & 24.351485 & 24.519494 & 24.104899 \\
\cline { 2 - 7 } & 3 & 15.53 & 24.636986 & 24.577900 & 24.657204 & 25.929835 \\
\cline { 2 - 7 } & 4 & 20.16 & 24.960974 & 24.751543 & 24.982344 & 25.065115 \\
\cline { 2 - 7 } & 5 & 18.94 & 25.448985 & 25.553804 & 24.860717 & 26.549726 \\
\hline \multirow{4}{*}{2} & 2 & 16.77 & 24.48804 & 24.584723 & 24.650583 & 24.747813 \\
\cline { 2 - 7 } & 3 & 18.85 & 24.866677 & 24.582997 & 24.559801 & 25.454637 \\
\cline { 2 - 7 } & 4 & 20.12 & 24.964328 & 24.782696 & 25.054277 & 25.260387 \\
\cline { 2 - 7 } & 5 & 21.71 & 26.76419 & 25.037394 & 25.151130 & 26.946907 \\
\hline \multirow{3}{*}{3} & 2 & 14.36 & 24.649292 & 24.477749 & 24.655701 & 25.288265 \\
\cline { 2 - 7 } & 3 & 19.53 & 24.829821 & 24.644722 & 24.769577 & 25.408871 \\
\cline { 2 - 7 } & 4 & 20.99 & 24.865398 & 24.870877 & 25.056663 & 25.196562 \\
\cline { 2 - 7 } & 5 & 20.54 & 25.093037 & 24.858501 & 25.064660 & 26.927766 \\
\hline
\end{array}
$$

Table 14 : Comparison between Modified Artificial Bee Colony algorithm (MABC), Artificial Bee Colony algorithm (ABC), Particle Swarm Optimization algorithm (PSO) and Genetic Algorithm (GA) algorithms using Tsallis entropy with proposed DE+OTSU based on SSIM \cite{25} values.

$$
\begin{array}{|l|l|l|c|c|c|c|}
\hline \multirow{2}{*}{\text { Image }} & \multirow{2}{*}{\text { Level }} & \multirow{2}{*}{\text { DE+OTSU }} & \multicolumn{4}{|c|}{\text { TSALLIS ENTROPY }} \\
\cline { 4 - 7 } & & & \text { MABC } & \text { ABC } & \text { PSO } & \text { GA } \\
\hline \multirow{3}{*}{1} & 2                      & 0.37                                         & 0.873413                 & 0.848868                & 0.904213                & 0.877822 \\
\cline { 2 - 7 } & 3                      & 0.312                                        & 0.938499                 & 0.927630                & 0.854607                & 0.922841 \\
\cline { 2 - 7 } & 4                      & 0.59                                         & 0.964424                 & 0.946915                & 0.962464                & 0.961415 \\
\cline { 2 - 7 } & 5                      & 0.492                                        & 0.979107                 & 0.978985                & 0.944744                & 0.964881 \\
\hline \multirow{4}{*}{2} & 2                      & 0.53                                         & 0.8708                   & 0.839311                & 0.888775                & 0.902111 \\
\cline { 2 - 7 } & 3                      & 0.59                                         & 0.93286                  & 0.847479                & 0.848088                & 0.945562 \\
\cline { 2 - 7 } & 4                      & 0.761                                        & 0.96772                  & 0.948136                & 0.931402                & 0.963708 \\
\cline { 2 - 7 } & 5                      & 0.672                                        & 0.978641                 & 0.969851                & 0.977266                & 0.924099 \\
\hline \multirow{3}{*}{3} & 2                      & 0.421                                        & 0.923064                 & 0.860053                & 0.906366                & 0.877859 \\
\cline { 2 - 7 } & 3                      & 0.672                                        & 0.939928                 & 0.915346                & 0.891213                & 0.816022 \\
\cline { 2 - 7 } & 4                      & 0.73                                         & 0.960756                 & 0.943489                & 0.942365                & 0.935665 \\
\cline { 2 - 7 } & 5                      & 0.704                                        & 0.968972                 & 0.953955                & 0.964825                & 0.967876 \\
\hline
\end{array}
$$

\section{CONCLUSIONS}
In this study, an algorithm for image segmentation task based on a combination of Differential evolution and Otsu’s multi-level thresholding was developed and the results were experimentally examined against three different satellite images. The study also explored the comparison between the proposed algorithm and the developed MABC\cite{11}, ABC, PSO, GA using Kapur’s entropy, Between-class variance and Tsallis entropy (one at a time) based on the three satellite images based on PSNR, SSIM and CPU time parameters.
The proposed Differential evolution based Otsu method(DE+OTSU) yields promising results which are close to the results obtained from the other algorithms (MABC, ABC and PSO) and we observed to be better than Genetic algorithm(GA). On proper parameter setting, there is a positive probability of increased performance results when using the proposed algorithm.

\begin{figure}[!htb]
\begin{subfigure}{\textwidth}
\includegraphics[width=\textwidth,trim={0 0 0 4.5cm}]{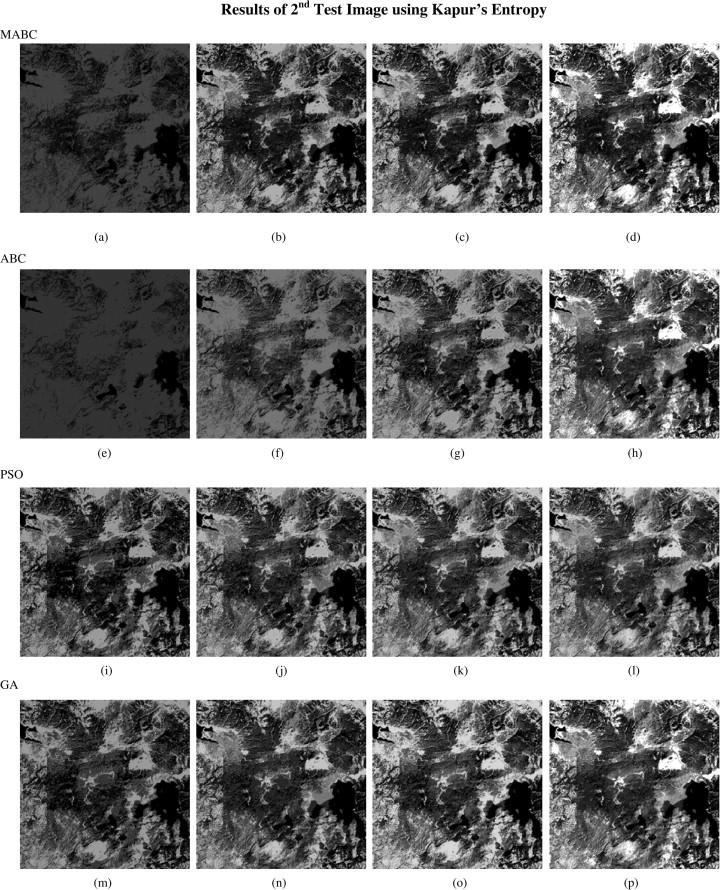}
\end{subfigure}
DE+OTSU
\newline

\begin{subfigure}{0.24\textwidth}
\includegraphics[width=\textwidth, scale=0.75]{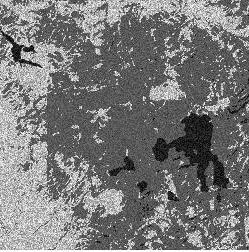}
\centering(q)
\end{subfigure}
\hfill
\begin{subfigure}{0.24\textwidth}
\includegraphics[width=\textwidth, scale=0.75]{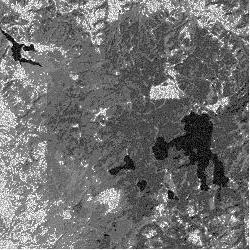}
\centering(r)
\end{subfigure}
\hfill
\begin{subfigure}{0.24\textwidth}
\includegraphics[width=\textwidth, scale=0.75]{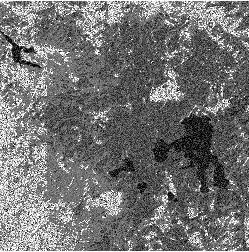}
\centering(s)
\end{subfigure}
\hfill
\begin{subfigure}{0.24\textwidth}
\includegraphics[width=\textwidth, scale=0.75]{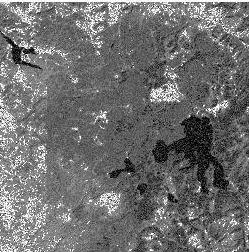}
\centering(t)
\end{subfigure}
\hfill
\label{fig:SatelliteImgs}
\caption{Segmented images obtained for 1st satellite image using MABC, ABC, PSO, GA with Kapur’s entropy and proposed DE+OSU methods. (a-d): levels 2-5 from MABC using Kapur’s entropy. (e-h): levels 2-5 from ABC using  Kapur’s entropy. (i-l): levels 2-5 from PSO using Kapur’s entropy. (m-p): levels 2-5 from GA using Kapur’s entropy. (q-t): levels 2-5 from proposed DE+OTSU algorithm.}
\end{figure}

\begin{figure}[!htb]
\begin{subfigure}{\textwidth}
\includegraphics[width=\textwidth,trim={0 0 0 4.5cm}]{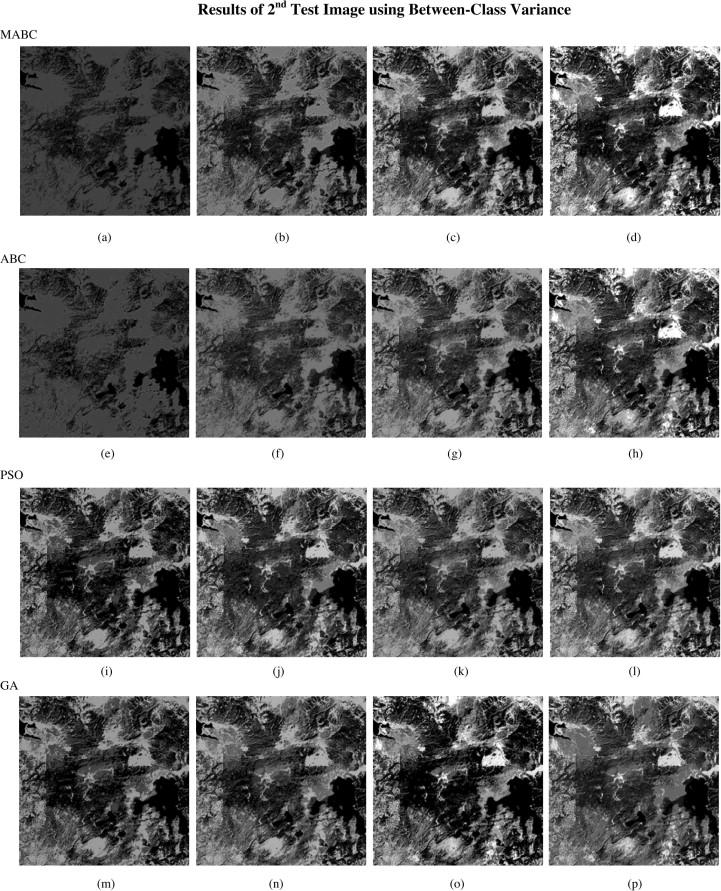}
\end{subfigure}
DE+OTSU
\newline

\begin{subfigure}{0.24\textwidth}
\includegraphics[width=\textwidth, scale=0.75]{img_1/im_1_4.jpg}
\centering(q)
\end{subfigure}
\hfill
\begin{subfigure}{0.24\textwidth}
\includegraphics[width=\textwidth, scale=0.75]{img_1/im_1_5.jpg}
\centering(r)
\end{subfigure}
\hfill
\begin{subfigure}{0.24\textwidth}
\includegraphics[width=\textwidth, scale=0.75]{img_1/im_1_6.jpg}
\centering(s)
\end{subfigure}
\hfill
\begin{subfigure}{0.24\textwidth}
\includegraphics[width=\textwidth, scale=0.75]{img_1/im_1_7.jpg}
\centering(t)
\end{subfigure}
\hfill
\label{fig:SatelliteImgs}
\caption{Segmented images obtained for 1st satellite image using MABC,ABC,PSO,GA with Between-class variance and proposed DE+OSU methods. (a-d) : levels 2-5 from MABC using Between-class variance. (e-h) : levels 2-5 from ABC using Between-class variance. (i-l) : levels 2-5 from PSO using Between-class variance.(m-p) : levels 2-5 from GA using Between-class variance. (q-t) : levels 2-5 from proposed DE+OTSU algorithm.}
\end{figure}

\begin{figure}[!htb]
\begin{subfigure}{\textwidth}
\includegraphics[width=\textwidth,trim={0 0 0 4.5cm}]{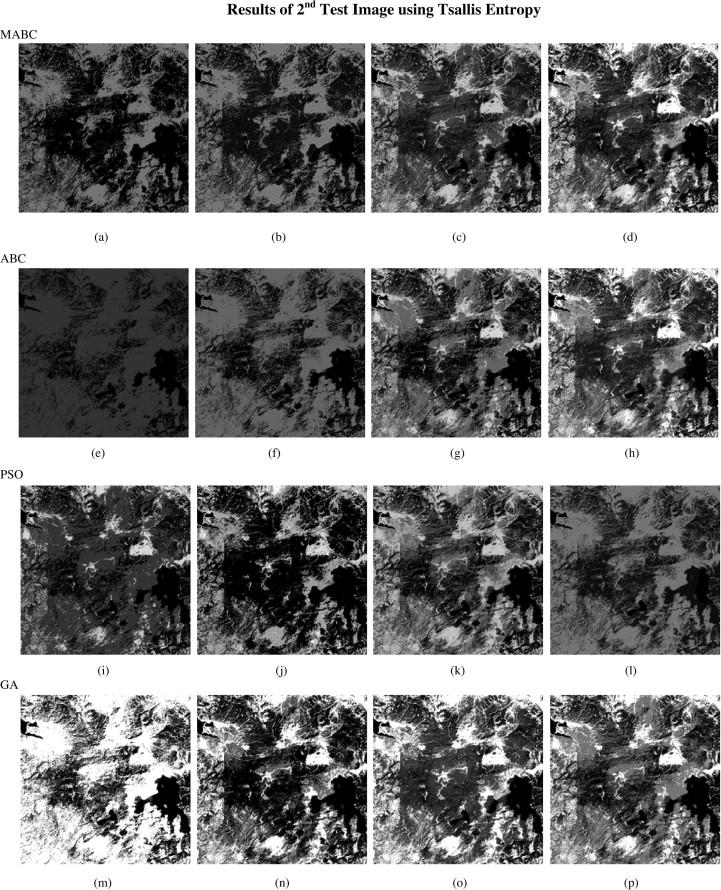}
\end{subfigure}
DE+OTSU
\newline

\begin{subfigure}{0.24\textwidth}
\includegraphics[width=\textwidth, scale=0.75]{img_1/im_1_4.jpg}
\centering(q)
\end{subfigure}
\hfill
\begin{subfigure}{0.24\textwidth}
\includegraphics[width=\textwidth, scale=0.75]{img_1/im_1_5.jpg}
\centering(r)
\end{subfigure}
\hfill
\begin{subfigure}{0.24\textwidth}
\includegraphics[width=\textwidth, scale=0.75]{img_1/im_1_6.jpg}
\centering(s)
\end{subfigure}
\hfill
\begin{subfigure}{0.24\textwidth}
\includegraphics[width=\textwidth, scale=0.75]{img_1/im_1_7.jpg}
\centering(t)
\end{subfigure}
\hfill
\label{fig:SatelliteImgs}
\caption{Segmented images obtained for 1st satellite image using MABC,ABC,PSO,GA with Tsallis entropy and proposed DE+OSU methods. (a-d) : levels 2-5 from MABC using Tsallis entropy. (e-h) : levels 2-5 from ABC using Tsallis entropy. (i-l) : levels 2-5 from PSO using Tsallis entropy.(m-p) : levels 2-5 from GA using Tsallis entropy. (q-t) : levels 2-5 from proposed DE+OTSU algorithm.}
\end{figure}

\begin{figure}[!htb]
\begin{subfigure}{\textwidth}
\includegraphics[width=\textwidth,trim={0 0 0 4.5cm}]{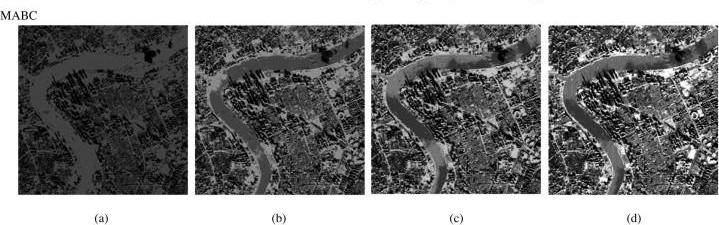}
\centering levels 2-5 from MABC using Kapur’s entropy.
\end{subfigure}

\begin{subfigure}{\textwidth}
\includegraphics[width=\textwidth, scale=0.75]{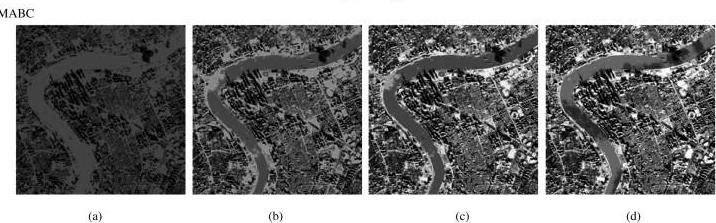}
\centering levels 2-5 from MABC using Between-Class variance.
\end{subfigure}

\begin{subfigure}{\textwidth}
\includegraphics[width=\textwidth, scale=0.75]{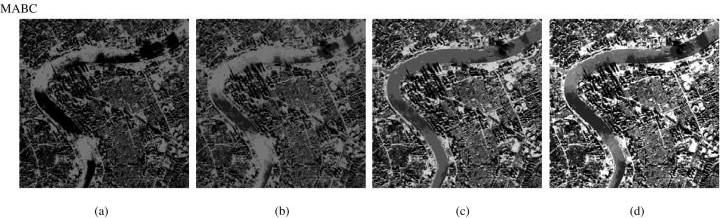}
\centering levels 2-5 from MABC using Tsallis entropy.\newline

\end{subfigure}

\begin{subfigure}{0.24\textwidth}
\includegraphics[width=\textwidth, scale=0.75]{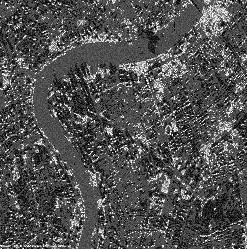}
\centering(a)
\end{subfigure}
\hfill
\begin{subfigure}{0.24\textwidth}
\includegraphics[width=\textwidth, scale=0.75]{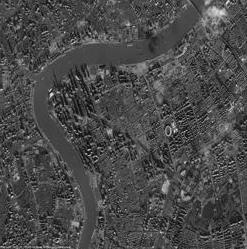}
\centering(b)
\end{subfigure}
\hfill
\begin{subfigure}{0.24\textwidth}
\includegraphics[width=\textwidth, scale=0.75]{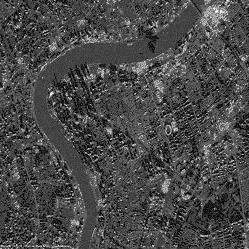}
\centering(c)
\end{subfigure}
\hfill
\begin{subfigure}{0.24\textwidth}
\includegraphics[width=\textwidth, scale=0.75]{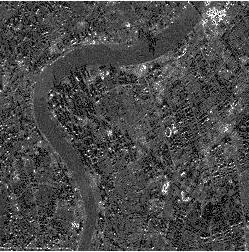}
\centering(d)
\end{subfigure}
\centering levels 2-5 from proposed DE+OTSU algorithm.

\label{fig:SatelliteImgs}
\caption{Segmented images obtained for 2nd satellite image using MABC and proposed DE+OSU methods.}
\end{figure}


\begin{figure}[!htb]
\begin{subfigure}{\textwidth}
\includegraphics[width=\textwidth,trim={0 0 0 4.5cm}]{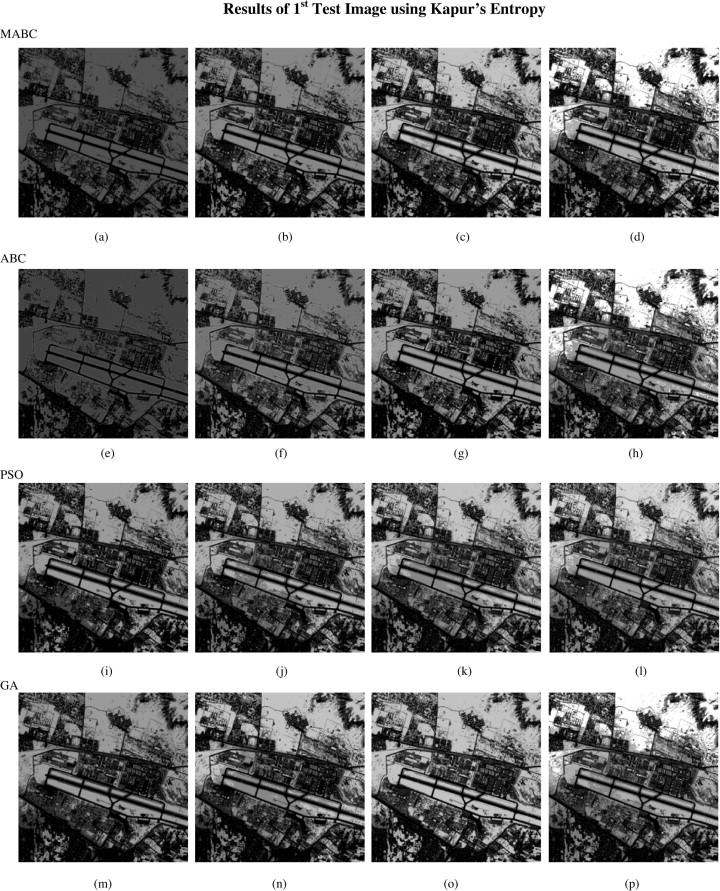}
\end{subfigure}
DE+OTSU
\newline
\begin{subfigure}{0.24\textwidth}
\includegraphics[width=\textwidth, scale=0.75]{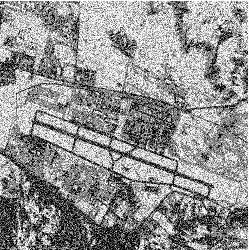}
\centering(q)
\end{subfigure}
\hfill
\begin{subfigure}{0.24\textwidth}
\includegraphics[width=\textwidth, scale=0.75]{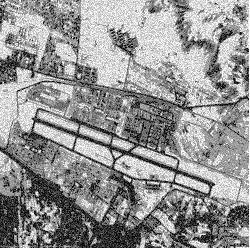}
\centering(r)
\end{subfigure}
\hfill
\begin{subfigure}{0.24\textwidth}
\includegraphics[width=\textwidth, scale=0.75]{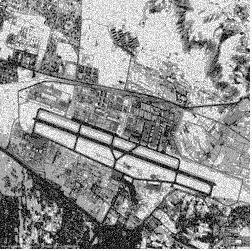}
\centering(s)
\end{subfigure}
\hfill
\begin{subfigure}{0.24\textwidth}
\includegraphics[width=\textwidth, scale=0.75]{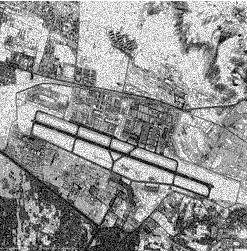}
\centering(t)
\end{subfigure}
\hfill
\label{fig:SatelliteImgs}
\caption{Segmented images obtained for 3rd satellite image using MABC,ABC,PSO,GA with Kapur’s entropy and proposed DE+OSU methods.(a-d) : levels 2-5 from MABC using Kapur’s entropy.(e-h) : levels 2-5 from ABC using Kapur’s entropy.(i-l) : levels 2-5 from PSO using Kapur’s entropy.(m-p) : levels 2-5 from GA using Kapur’s entropy.(q-t) : levels 2-5 from proposed DE+OTSU algorithm.}
\end{figure}

\begin{figure}[!htb]
\begin{subfigure}{\textwidth}
\includegraphics[width=\textwidth,trim={0 0 0 4.5cm}]{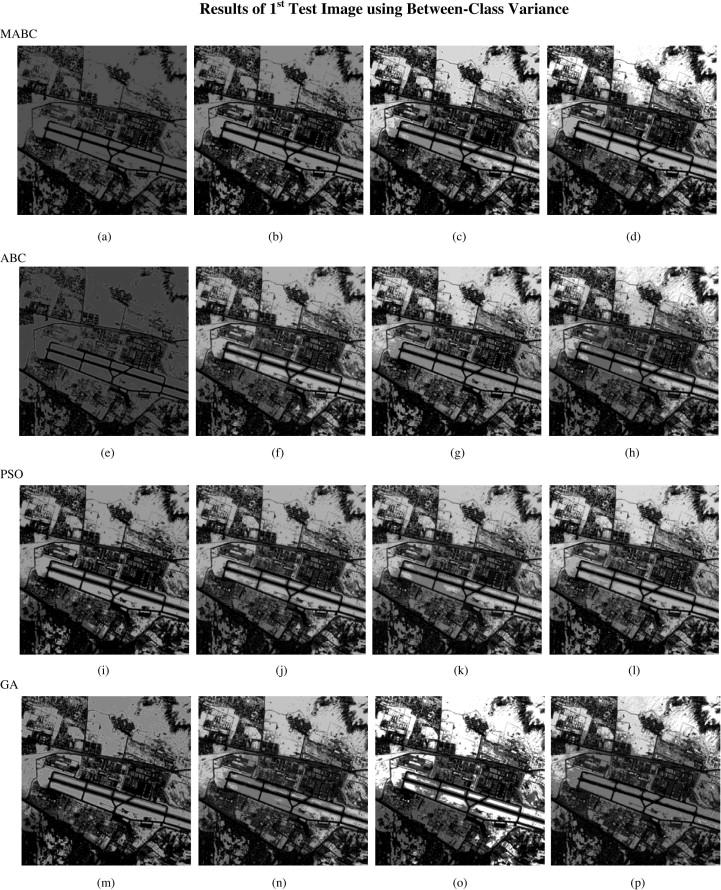}
\end{subfigure}
DE+OTSU
\newline
\begin{subfigure}{0.24\textwidth}
\includegraphics[width=\textwidth, scale=0.75]{img_3/im_3_4.jpg}
\centering(q)
\end{subfigure}
\hfill
\begin{subfigure}{0.24\textwidth}
\includegraphics[width=\textwidth, scale=0.75]{img_3/im_3_5.jpg}
\centering(r)
\end{subfigure}
\hfill
\begin{subfigure}{0.24\textwidth}
\includegraphics[width=\textwidth, scale=0.75]{img_3/im_3_6.jpg}
\centering(s)
\end{subfigure}
\hfill
\begin{subfigure}{0.24\textwidth}
\includegraphics[width=\textwidth, scale=0.75]{img_3/im_3_7.jpg}
\centering(t)
\end{subfigure}
\hfill
\label{fig:SatelliteImgs}
\caption{Segmented images obtained for 3rd satellite image using MABC,ABC,PSO,GA with Between-class variance and proposed DE+OSU methods. (a-d) : levels 2-5 from MABC using Between-class variance. (e-h) : levels 2-5 from ABC using Between-class variance. (i-l) : levels 2-5 from PSO using Between-class variance.(m-p) : levels 2-5 from GA using Between-class variance. (q-t) : levels 2-5 from proposed DE+OTSU algorithm.}
\end{figure}

\begin{figure}[!htb]
\begin{subfigure}{\textwidth}
\includegraphics[width=\textwidth,trim={0 0 0 4.5cm}]{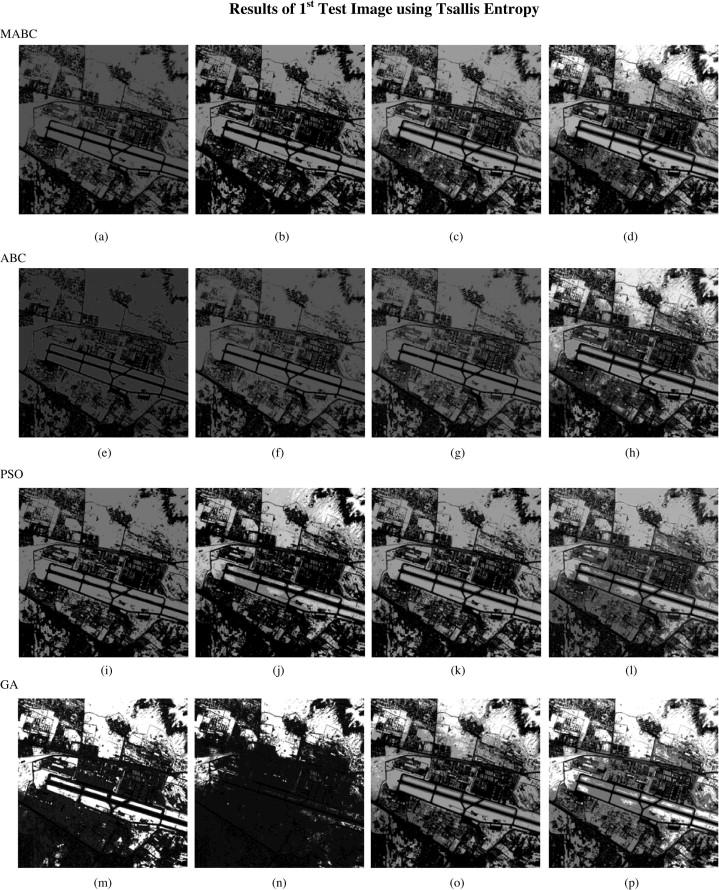}
\end{subfigure}
DE+OTSU
\newline
\begin{subfigure}{0.24\textwidth}
\includegraphics[width=\textwidth, scale=0.75]{img_3/im_3_4.jpg}
\centering(q)
\end{subfigure}
\hfill
\begin{subfigure}{0.24\textwidth}
\includegraphics[width=\textwidth, scale=0.75]{img_3/im_3_5.jpg}
\centering(r)
\end{subfigure}
\hfill
\begin{subfigure}{0.24\textwidth}
\includegraphics[width=\textwidth, scale=0.75]{img_3/im_3_6.jpg}
\centering(s)
\end{subfigure}
\hfill
\begin{subfigure}{0.24\textwidth}
\includegraphics[width=\textwidth, scale=0.75]{img_3/im_3_7.jpg}
\centering(t)
\end{subfigure}
\hfill
\label{fig:SatelliteImgs}
\caption{Segmented images obtained for 3rd satellite image using MABC,ABC,PSO,GA with Tsallis entopy and proposed DE+OSU methods. (a-d) : levels 2-5 from MABC using Tsallis entopy. (e-h) : levels 2-5 from ABC using Tsallis entopy. (i-l) : levels 2-5 from PSO using Tsallis entopy.(m-p) : levels 2-5 from GA using Tsallis entopy. (q-t) : levels 2-5 from proposed DE+OTSU algorithm.}
\end{figure}
\bibliographystyle{unsrt}  

\end{document}